\documentclass[11pt]{article}
\usepackage[left=2in,right=2in,verbose=true,letterpaper]{geometry}
\usepackage[table,xcdraw]{xcolor}
\usepackage{arxiv}
\usepackage{hyperref}
\usepackage{todonotes}
\usepackage{enumitem}

\usepackage[utf8]{inputenc} 
\usepackage[T1]{fontenc}    
\usepackage{hyperref}       
\usepackage{url}            
\usepackage{booktabs}       
\usepackage{amsfonts}       
\usepackage{nicefrac}       
\usepackage{microtype}      
\usepackage{lipsum}
\usepackage{footnote}
\usepackage{booktabs}
\makesavenoteenv{table}
\usepackage{comment}

\title{TorchKGE: \\Knowledge Graph Embedding in Python and PyTorch}

\author{
  Armand Boschin\\
  Télécom Paris\\
  Institut Polytechnique de Paris\\
  Paris, France\\
  \texttt{aboschin@enst.fr} \\
}

\begin{document}
\maketitle

\begin{abstract}
    TorchKGE is a Python module for knowledge graph (KG) embedding relying solely on PyTorch. This package provides researchers and engineers with a clean and efficient API to design and test new models. It features a KG data structure, simple model interfaces and modules for negative sampling and model evaluation. Its main strength is a very fast evaluation module for the link prediction task, a central application of KG embedding. Various KG embedding models are also already implemented. Special attention has been paid to code efficiency and simplicity, documentation and API consistency. It is distributed using PyPI under BSD license. Source code and pointers to documentation and deployment can be found at \url{https://github.com/torchkge-team/torchkge}.
\end{abstract}


\section{Introduction}
\label{intro}

A knowledge graph (KG) is a database consisting of a set of entities and a set of facts linking those entities. A fact is a (\textit{head, relation, tail}) triplet linking two entities by a typed \textit{relation}. A KG embedding is a vectorial representation of the graph; entities are usually represented as vectors while relations can either be vectors or bilinear transformations in the vector space. Generally, embeddings are used to quantify similarity between objects through distances in the vector space. In the case of KGs, embeddings are used to quantify the likelihood of any triplet, so as to predict new links and to classify triplets.

Although most authors of KG embedding recent work use common evaluation procedures (link prediction and triplet classification), it is often difficult to compare models in the same setting, in terms of programming language and hardware configuration. Indeed the authors' code is not always provided and when it is, a variety of programming languages (e.g., C++, Python, Java) and computing frameworks (e.g., Theano, SciPy, Tensorflow, PyTorch) are used. Papers such as \cite{Ruffinelli2020You} show that some recent advances are more likely to be caused by better training procedures than by better models. That shows the importance of a common training framework for model comparison.

It is important to note that all authors compare their models on link-prediction tasks, which measure how well the model can complete facts. For each fact ($h, r, t$) in a KG, a head test (resp. tail test) is done to rank all entities $e$ by decreasing order of score for the fact $(e, r, t)$ (resp. $(h, r, e)$). This rankings give the recovery ranks of true $h$ and $t$ and three standard metrics can be reported: mean rank (MR), mean reciprocal rank (MRR) and hit-at-k (Hit@k). It is crucial to do this evaluation step very efficiently, in order to be able to do it as often as possible. For example, frequent evaluation during the training process allows a better fine-tuning of the parameters.

We propose here an efficient API for the implementation and evaluation of KG embedding models, including some state-of-the-art model implementations. This new open-source library is called TorchKGE for Knowledge Graph Embedding in PyTorch. It is currently in version 0.16.6.

\section{Related work}

There already exists several Python libraries for KG embedding. The most important are detailed hereafter.
 
\label{related}
\begin{itemize}[leftmargin=*]
    \setlength\itemsep{0em}
	\item OpenKE. First released in 2017\cite{han_openke:_2018}, it was first based on Tensorflow but PyTorch was recently adopted as the main computing framework and some features are implemented in C++. In spite of its good performance in training, evaluation is quite slow (see Section~\ref{perf}). It is also not included in any package manager and has no versioning, which makes it quite difficult to use.
	\item AmpliGraph. First released in March 2019 and currently in version 1.3.1, it is based on TensorFlow \cite{ampligraph}. It is efficient for models training but slow for link-prediction evaluation (see Section~\ref{perf}). It is also well documented.
	\item Pykg2vec. First released in March 2019 \cite{pykg2vec} and currently in version 0.0.50, it is based on TensorFlow. It aims to provide a wide variety of features (models, wrappers, visualization), but its development is still at a very early-stage and the API is not stable. We were unable to make it work with enough satisfaction to compare it to the others.
\end{itemize}

PyTorch-BigGraph \cite{lerer_pytorch-biggraph:_2019} is also worth mentioning for massive knowledge graph embedding though it is not the same use-case as the one at hand in this paper. OpenKE and AmpliGraph seem to be the two best candidates for providing a simple and unified API for KG embedding. Their implementations of link-prediction evaluation, which is a core application of KG embedding, is highly inefficient. We propose with TorchKGE a new API, which makes fast link-prediction evaluation core objective.

\section{Project vision}
\noindent \textbf{\textit{Goal.}} TorchKGE provides users with a KG data-structure, fast evaluation modules, off-the-shelf model implementations and model interfaces making it easy to design and test new architectures.

\noindent \textbf{\textit{PyTorch.}} PyTorch is an open-source machine learning library developed by FAIR under BSD license \cite{NEURIPS2019_9015}. It provides support of GPU acceleration for tensor computations along with an auto-differentiation system useful to train models with iterative methods. It is simple to learn and use and yet highly efficient. It is also seamlessly integrated with the Numpy library \cite{doi:10.1109/MCSE.2011.37}, which is one of the most widespread tools in machine learning research. This makes PyTorch one of the best choice to build a KG embedding framework.

\noindent \textbf{\textit{Design.}} The aim is to align the design of the API with that of PyTorch. Models are defined as instances of \texttt{torch.nn.module} without further wrapping, giving the user flexibility but still taking advantage of the power of PyTorch (GPU migration, optimization algorithms). The library includes as few dependencies as possible.

\noindent \textbf{\textit{Documentation.}} \textit{TorchKGE} provides a complete documentation, which is accessible online (\url{https://torchkge.readthedocs.io/en/latest/}). It includes class and method descriptions along with example scripts for model definition, training and evaluation. All existing model implementations include references to the original papers.

\noindent \textbf{\textit{Community.}} The project is developed using \textit{git} DVCS and a public mirror is available on \textit{GitHub}. External contributions are strongly encouraged when respecting the guidelines defined on the repository and in the documentation.

\noindent \textbf{\textit{Deployment.}} Installing and using TorchKGE is easy as the library is proposed on PyPI, one of the main package manager in Python. It is proposed under BSD license.

\section{Code structure}

\subsection{Models}
Model classes are implementations of PyTorch’s \verb|torch.nn.Module| in order to make use of the auto-grad feature seamlessly on the parameters. Each model class implements the \verb|models.interfaces.Models| interface. It requires the definition of four methods: \verb|scoring_function| (compute the score for each fact of a batch), \verb|normalize_parameters|, \verb|lp_scoring_function| (compute the score of all possible candidate triplets for each fact of a batch), \verb|lp_prep_cands| (preprocess embeddings and candidate entities for \verb|lp_scoring_function|).

\noindent In its current version, TorchKGE features five translation models (TransE, TransH, TransR, TransD and TorusE), 
five bilinear models (RESCAL, DistMult, HolE, ComplEx and ANALOGY), 
and one deep model (ConvKB) 
(See \cite{suchanek_knowledge_2019} and references therein). Pre-trained versions for off-the-shelf usage are continuously added.

\subsection{Model evaluation}
There are two evaluation modules:

\noindent \textbf{Link Prediction.} Evaluate the quality of the embedding by measuring how well the model can complete facts. As explained in Section~\ref{intro}, it is necessary to compute the scores of all candidate facts for each head and tail tests associated to the facts of the KG at hand. Once the recovery ranks of the true heads and tails are computed, the module can return the three standard metrics (MR, MRR, Hit@k) in raw and filtered settings \cite{suchanek_knowledge_2019}.

This evaluation process is done very efficiently thanks to the forced definition in the model of the \verb|lp_scoring_function| method. While most frameworks loop over the facts of a KG to compute the scores of related candidate facts, this method makes it possible to group those computations by batch for parallel computation, dramatically increasing the speed. It is also further-accelerated by pre-computing the model specific projections when possible (e.g. TransH). See Section~\ref{perf} for details on the performance of this module.

\noindent \textbf{Triplet Classification.} Evaluate the quality of the embedding by measuring how well the model classifies triplets as right or wrong. For initialization, the evaluator needs a trained model, a validation and a test set. On the validation set, score thresholds are computed for each relation so that any fact with a score higher than the threshold specific to the relation involved should be true. The thresholds are then used to classify as true or false the triplets of the test set and its negatively sampled version. The metric reported is the accuracy.

\subsection{Knowledge graphs in memory}
TorchKGE requires a representation of knowledge graphs in memory. This is the purpose of the \verb|data_structures.KnowledgeGraph| class. It features a \verb|split_kg| method to split the facts of the knowledge graph in train, test and optionally validation sets. When the split is random, the method keeps at least one fact involving each entity and each relation in the training subset. The \verb|utils.datasets| module provides functions to easily load common datasets such as FB13, FB15k, FB15k-237, WN18, WN18RR, YAGO3-10 and WikiDataSets (See \cite{suchanek_knowledge_2019} and references therein). The \verb|utils.data_redundancy| provides tools to measure redundancy in a dataset \cite{10.1145/3318464.3380599}.

\subsection{Negative sampling}
Negative sampling is the randomized generation of false facts. It is instrumental in the generation of relevant embeddings. TorchKGE implements various negative sampling methods as different classes implementing the interface \verb|sampling.NegativeSampler|. This interface requires its child classes to implement the \verb|corrupt_batch| method, which corrupts the batch of facts passed as arguments and that will be used during training. It also helps define the \verb|corrupt_kg| method, which can be used to corrupt all the facts of a KG at once. In the current version, TorchKGE features three methods in the \verb|sampling| module: uniform, Bernoulli and positional negative sampling (See \cite{suchanek_knowledge_2019} and references therein).

\section{Performance}
\label{perf}

OpenKE (April 9, 2020 version), AmpliGraph (v1.3.1) and TorchKGE (v0.16.0) were compared in terms of running times. We first aimed to include pykg2vec (v0.50.0) in the comparison but we did not manage to make it work sufficiently well, due to its unstable API and outdated documentation. Four different models (TransE, TransD, RESCAL and ComplEx) were trained in each framework (apart from AmpliGraph which does not feature TransD and RESCAL) on both FB15k and WN18 and then evaluated by link prediction on test sets. The goal was to measure the mean epoch time and evaluation duration in each case. 

Experiments were done TensorFlow 1.15 and a Tesla K80 GPU powered by Cuda 10.2. For the CPU parallelization of OpenKE, 8 threads were used. For comparison purpose, the same sets of hyper-parameters were used in each framework: Adam optimizer, 0.01 as learning rate, 1e-5 as L2-regularization factor, Bernoulli negative sampling and one negative sample per triplet. Model specific hyper-parameters are detailed in Table~\ref{tab:params}.

In Table~\ref{tab:training}, the mean durations of the first 100 epochs of training are reported. They were averaged over 10 independent processes though the mean duration is quite stable (the standard deviations were not significant). We note that TorchKGE's mean epoch times are of the same order as those of the other frameworks, and always within a factor 2.

In Table~\ref{tab:eval}, the duration of the link-prediction evaluation process is reported for each model in each framework. These were averaged over 5 independent processes but standard deviations were not significant. We note that TorchKGE is significantly faster than the two other frameworks; always at least three times faster and up to twenty-four times in the case of OpenKE when evaluating RESCAL on WN18.


\begin{table}
\centering
\begin{tabular}{l|c|c|c|c|}
\cline{2-5}
                                                             & TransE                   & TransD & RESCAL                & ComplEx               \\ \hline
\multicolumn{1}{|l|}{Number of batches}                      & \multicolumn{2}{c|}{10}           & 20                    & 10                    \\ \hline
\multicolumn{1}{|l|}{Embedding dimension}                    & \multicolumn{4}{c|}{100}                                                          \\ \hline
\multicolumn{1}{|l|}{Hidden dimension}                       & \cellcolor[HTML]{EFEFEF} & 50     & \multicolumn{2}{c|}{\cellcolor[HTML]{EFEFEF}} \\ \hline
\multicolumn{1}{|l|}{Loss}                                   & \multicolumn{2}{c|}{Margin loss}  & \multicolumn{2}{c|}{Sigmoid loss}             \\ \hline
\multicolumn{1}{|l|}{Margin}                                 & \multicolumn{2}{c|}{1}            & \multicolumn{2}{c|}{\cellcolor[HTML]{EFEFEF}} \\ \hline
\end{tabular}
\caption{Hyper-parameters used to train the models in all frameworks of the experiment.}
\label{tab:params}
\end{table}

\begin{table}[!ht]
\centering
\begin{tabular}{c|c|c|c|c|c|c|c|c|}
\cline{2-9}
                                 & \multicolumn{2}{l|}{TransE} &\multicolumn{2}{l|}{TransD} & \multicolumn{2}{l|}{RESCAL} & \multicolumn{2}{l|}{ComplEx} \\ \cline{2-9} 
                                 & FB15k & WN18 & FB15k & WN18 & FB15k & WN18 & FB15k & WN18 \\ \hline
\multicolumn{1}{|l|}{AmpliGraph} & \textbf{0.171} & \textbf{0.106} & \multicolumn{4}{c|}{\cellcolor[HTML]{EFEFEF}} & 0.528 & 0.254 \\ \hline
\multicolumn{1}{|l|}{OpenKE}     & 0.415 & 0.150 & \textbf{0.538} &\textbf{ 0.284} & \textbf{3.46} & \textbf{1.05} & 0.457 & \textbf{0.191} \\ \hline
\multicolumn{1}{|l|}{TorchKGE}   & 0.312 & 0.156 & 0.770 & 0.543 & 3.91 & 1.20 & \textbf{0.449} & 0.287 \\ \hline
\end{tabular}
\caption{Mean epoch duration (in seconds) of 10 independent training processes. Standard deviations were all smaller than 0.006, which is not significant compared to the reported values. Best values are in bold.}
\label{tab:training}
\end{table}

\begin{table}[!ht]
\centering
\begin{tabular}{l|c|c|c|c|c|c|c|c|}
\cline{2-9}
                                 & \multicolumn{2}{l|}{TransE} &\multicolumn{2}{l|}{TransD} & \multicolumn{2}{l|}{RESCAL} & \multicolumn{2}{l|}{ComplEx} \\ \cline{2-9} 
                                 & FB15k & WN18 & FB15k & WN18 & FB15k & WN18 & FB15k & WN18 \\ \hline
\multicolumn{1}{|l|}{AmpliGraph} & 354.8 & 39.8 & \multicolumn{4}{c|}{\cellcolor[HTML]{EFEFEF}} & 537.2 & 94.9 \\ \hline
\multicolumn{1}{|l|}{OpenKE}     & 235.6 & 42.2 & 258.5 & 43.7 & 789.1 & 178.4 & 354.7 & 63.9 \\ \hline
\multicolumn{1}{|l|}{TorchKGE}   & \textbf{76.1} & \textbf{13.8} & \textbf{60.8} & \textbf{11.1} & \textbf{46.9} & \textbf{7.06} & \textbf{96.4} & \textbf{18.6} \\ \hline
\end{tabular}
\caption{Duration (in seconds) of the link-prediction evaluation process on test sets. Those figures are averaged over 5 independent evaluation processes (standard deviation was never significant). Best values are in bold.}
\label{tab:eval}
\end{table}

\section{Conclusion}
Built upon the well-known PyTorch framework, TorchKGE already provides remarkable performance on various evaluation tasks. We believe that KG embedding research can greatly benefit from having an efficient and unified computing framework to improve the diffusion of knowledge.

\bibliographystyle{acm}
\bibliography{references}
\end{document}